\documentclass[letterpaper]{article}
\usepackage{aaai24}
\usepackage{times}
\usepackage{helvet}
\usepackage{courier}
\usepackage[hyphens]{url}
\usepackage{graphicx}
\usepackage{natbib}  
\usepackage{caption} 
\usepackage{algorithm}
\usepackage{algorithmic}

\usepackage{newfloat}
\usepackage{listings}
\DeclareCaptionStyle{ruled}{labelfont=normalfont,labelsep=colon,strut=off} 
\floatstyle{ruled}
\newfloat{listing}{tb}{lst}{}
\floatname{listing}{Listing}

\usepackage{booktabs} 
\usepackage{amsmath}
\usepackage{mathtools}
\usepackage{amssymb}

\title{AesFA: An Aesthetic Feature-Aware Arbitrary Neural Style Transfer}
\author{
    Joonwoo Kwon\textsuperscript{\rm 1}\equalcontrib,
    Sooyoung Kim\textsuperscript{\rm 1}\equalcontrib,
    Yuewei Lin\textsuperscript{\rm 2} \textsuperscript{\textdagger},
    Shinjae Yoo\textsuperscript{\rm 2} \textsuperscript{\textdagger},
    Jiook Cha\textsuperscript{\rm 1} \thanks{Co-corresponding authors.} \\
}

\affiliations{
    \textsuperscript{\rm 1}Seoul National University\\
    \textsuperscript{\rm 2}Brookhaven National Laboratory\\
    \{pioneers, rlatndud0513, connectome\}@snu.ac.kr, \; \{ywlin, sjyoo\}@bnl.gov
    }

\begin{document}
\maketitle


\begin{abstract}
Neural style transfer (NST) has evolved significantly in recent years. Yet, despite its rapid progress and advancement, existing NST methods either struggle to transfer aesthetic information from a style effectively or suffer from high computational costs and inefficiencies in feature disentanglement due to using pre-trained models. This work proposes a lightweight but effective model, {\bfseries AesFA}---{\bfseries Aes}thetic {\bfseries F}eature-{\bfseries A}ware NST. The primary idea is to decompose the image via its frequencies to better disentangle aesthetic styles from the reference image while training the entire model in an end-to-end manner to exclude pre-trained models at inference completely. To improve the network's ability to extract more distinct representations and further enhance the stylization quality, this work introduces a new aesthetic feature: contrastive loss. Extensive experiments and ablations show the approach not only outperforms recent NST methods in terms of stylization quality, but it also achieves faster inference. Codes are available at https://github.com/Sooyyoungg/AesFA.
\end{abstract}


\section{Introduction} \label{sec:introduction}
\begin{figure}[t!]
    \centering
    \includegraphics[width=0.48\textwidth, height=4.5cm]{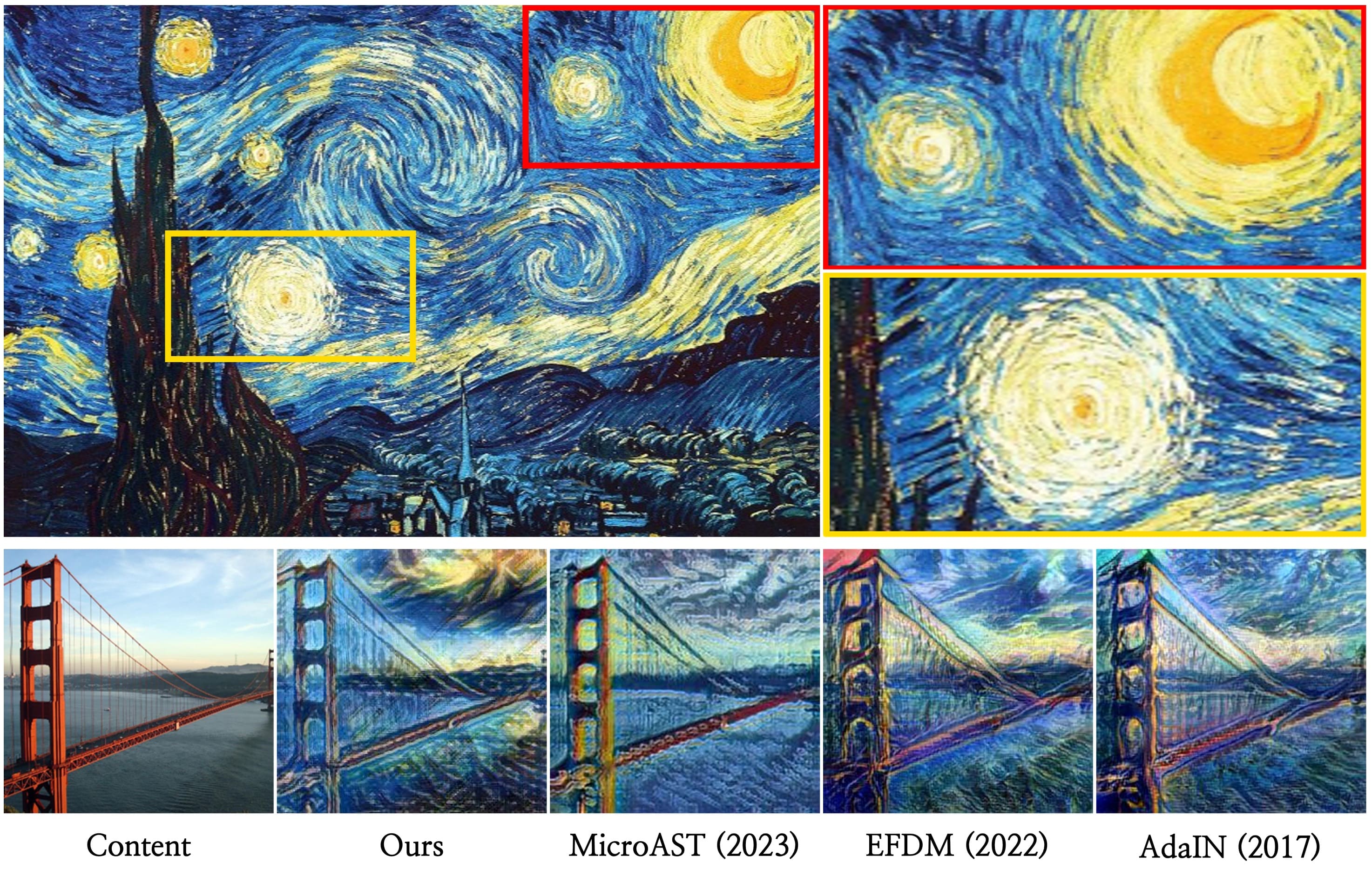} 
    \caption{Top: \textit{The Starry Night} by Vincent Van Gogh. The styles have a strong correlation with spatial information, as evidenced by the presence of whirling patterns and expressionistic yellow stars in the ``sky". Bottom: Compared with other NST methods, our method can faithfully transfer styles while ensuring the spatial information.} 
    \label{fig:fig1}
    \end{figure}

Neural Style Transfer (NST) is an artistic application that transfers the style of one image to another while preserving the original content. Initially introduced by~\cite{gatys2016image}, this area has gained substantial momentum with the advancement of deep neural networks. Despite such progress, a significant chasm persists between authentic artwork and synthesized stylizations. Existing NST methods, as shown in Figure~\ref{fig:fig1}, struggle to capture essential aesthetic features, such as tones, brushstrokes, textures, grains, and the local structure from style images, leading to discordant colors and irrelevant patterns. Ideally, the goal of using NST is to extract a style from the image and transfer it to content, necessitating representations that capture both image semantics and stylistic changes. This work focuses on defining these \textit{style} representations.

In the context of painting, style representations are defined by attributes, such as overall color and/or the local structure of brushstrokes. Most NST algorithms define style representations as spatially agnostic features to encode this information. For example, Gatys et al.~\cite{gatys2016image} use gram matrices, while Huang et al.~\cite{huang2017arbitrary} employ mean and variance alignment to obtain a style representation. Despite their success, they rely solely on summary statistics. Thus, they lack spatial information representation. In fact, style representations are \textit{highly correlated} to spatial information. For example, Vincent van Gogh's \textit{The Starry Night} (Figure~\ref{fig:fig1}) has expressionistic yellow stars and a moon that dominate the upper center and right, while dynamic swirls fill the center of the sky. In pondering the \textit{style} of this painting, its focal point primarily resides in the sky rather than the village or cypress trees. Therefore, when transferring \textit{The Starry Night}'s style, the expected style output likely would be the dynamic swirls and expressionistic yellow stars in the sky. From this point of view, spatial information keenly matters in style representations. However, most NST algorithms fail to recognize such distinct spatial styles due to their spatial-independent style representations, leading to stylizations lacking in spatial coherence (refer to the bottom panel in Figure~\ref{fig:fig1}). 

To enhance stylization, we propose a lightweight yet effective model that we call, \textbf{Aes}thetic \textbf{F}eature-\textbf{A}ware Arbitrary NST, or \textbf{AesFA}. AesFA overcomes prior NST limitations by encoding style representations while retaining spatial details. To expedite the extraction of aesthetic features, we decompose the image into two discrete complementary components, i.e., the high- and low-frequency parts. High frequency captures details including textures, grains, and brushstrokes, while low frequency encodes global structures and tones. On the other hand, existing NST algorithms often neglect this disentanglement and extract style features from a mix of irrelevant information. Specifically, we employ Octave Convolution operators (OctConv)~\cite{chen2019drop} to decompose and process input images by frequency, which eliminates the need for cumbersome mathematical algorithms like Fast Fourier Transform (FFT)~\cite{gentleman1966fast}. This design ensures the model remains lightweight and effective when disentangling features. Furthermore, inspired by adaptive convolutions (AdaConv)~\cite{chandran2021adaptive}, which simultaneously blend statistical and structural styles to the contents, we modify AdaConv by effectively combining frequency-decomposed content features with predicted \textit{aesthetic feature-aware kernels and biases}. We refer to the modified stylization module as \textit{Adaptive Octave Convolution (AdaOct)} because it employs AdaConv followed by an OctConv. In AdaOct, frequency-decomposed features undergo convolution with predicted aesthetic-aware kernels and biases, followed by OctConv for exchanging features' high- and low-frequency components. Consequently, AdaOct achieves superior stylization and reduces unwanted artifacts.

Another challenge is that existing NST methods heavily rely on pre-trained networks, e.g., VGG~\cite{simonyan2014very}, for feature extraction. However, using such networks during inference is inefficient because of the computational demands from fully connected layers. This limits NST's use at high resolutions (e.g., 2K; 4K) and in mobile or real-time scenarios. Larger images also struggle with preserving texture and grains in style transfer. To mitigate this limitation, a prior study \cite{wang2023microast} adopted contrastive learning for end-to-end training while excluding pre-trained convolutional neural networks (CNNs) at inference. However, this approach is computationally expensive and inefficient as it uses all negative samples in a mini-batch, especially with higher-resolution samples. This prompts a question: \textit{are all negative samples necessary?} Intuitively, the more distant negative samples contribute less to training as they are already well discriminated from the positive sample and vice versa. Inspired by hard negative mining, we redefine ``negative" samples as the \textit{k}-th nearest negative samples to the stylized output, introducing efficient contrastive learning for aesthetic features via pre-trained VGG network.

Overall, AesFA outperforms state-of-the-art algorithms in terms of the structural similarity index (SSIM) and average VGG style perceptual loss across all spatial resolutions, ranging from 256 to 4K. Regardless of image resolution, our method achieves state-of-the-art performance, inferring a single image in under 0.02 seconds.

The contributions of this work are summarized as follows:
\begin{itemize}
\item We propose a lightweight yet effective model for aesthetic feature-aware NST, which maintains the spatial style information and decomposes images by frequency to improve feature extraction, substantially enhancing the stylization quality and computational efficiency at the same time. 
\item To effectively infuse frequency-decomposed content features with aesthetic features, a new stylization module, AdaOct, is proposed that yields more satisfying stylizations with sophisticated aesthetic characteristics. To further accelerate the networks' capability to extract more distinct aesthetic representations, a straightforward contrastive learning for aesthetic features also is proposed.
\item We show that our method achieves generalization, quality, and efficiency simultaneously across various spatial resolutions by conducting comprehensive comparisons with several state-of-the-art NST methods.
\end{itemize}


\section{Related Work} \label{sec:Relatedwork}
\textbf{NST} emerged with Gatys et al.~\cite{gatys2016image}, but its optimization is computationally intensive. To deal with this issue, Johnson et al.~\cite{johnson2016perceptual} introduced perceptual losses for real-time processing. Subsequent work~\cite{gatys2017controlling, ghiasi2017exploring, chen2016fast, ulyanov2017improved, dumoulin2016learned, ulyanov2016texture} improved NST without sacrificing speed. However, all were limited to specific styles. Huang et al.~\cite{huang2017arbitrary} proposed Adaptive Instance Normalization (AdaIN) for arbitrary style transfer, which has been extended~\cite{sheng2018avatar,kotovenko2019content, jing2020dynamic, shen2018neural, li2017universal, lin2021drafting, yoo2019photorealistic} for successful style transfer onto content images. Chandran et al.~\cite{chandran2021adaptive} improved AdaIN with AdaConv for structure-aware style transfer. AdaConv simultaneously adapts statistical and structural styles. However, AdaConv's convolution kernels and biases incur high computational costs. Recent work~\cite{an2023bigger, wang2023microast, wang2021rethinking} has highlighted drawbacks in NST methods that rely on pre-trained CNNs, e.g., VGG-19~\cite{simonyan2014very}, for feature extraction from the reference image. Wang et al.~\cite{wang2021rethinking} have enhanced non-VGG architectures' robustness via activation smoothing in stylization loss. An et al.~\cite{an2023bigger} explore alternative architectures, such as GoogLeNet~\cite{szegedy2015going}, yet they lack specificity for NST, yielding unsatisfactory stylization outcomes and high memory use. Instead, our objective is productive mobile NST that incorporates aesthetic features.

\textbf{Multiscale representation learning.} Prior to the advent of deep learning, multiscale representation, such as scale-invariant feature transform (SIFT) features \cite{lowe2004distinctive}, was used for local feature extraction. It remains valuable for its robustness and generalization in the deep learning era. Methods like FPN (Feature Pyramid Network)~\cite{lin2017feature} and PSP (Pyramid Scene Parsing Network)~\cite{zhao2017pyramid} combine convolutional features for object detection and segmentation. Meanwhile, network architectures, e.g., \cite{chen2018big, sun2019deep, wang2019elastic, huang2017multi, ke2017multigrid}, exploit multiscale features effectively. Enhanced designs like OctConv \cite{chen2019drop} exchange inter-frequency information, reducing redundancy and improving CNN classification. Multiscale representation's prowess is harnessed across various vision tasks: image classification \cite{wang2021dual}, compression \cite{akbari2020generalized, liu2021image}, enhancement \cite{huo2021efficient, li2020frequency, zhang2022multi}, and generation \cite{wang2020semi, durall2019stabilizing}. Our model, AesFA, decomposes input images by frequencies to extract and transfer aesthetic features from style images, reducing computational costs.

\textbf{Frequency analysis in deep learning.} Traditional image processing (\cite{van1992computational, johnson2006modified}) has extensively used frequency analysis. Studies connect frequency analysis with deep learning techniques \cite{chen2019drop, xu2020learning, xu2019frequency, durall2020watch}. Wang et al.~\cite{wang2020high} highlight high-frequency components' role in neural networks' generalization. Czoble et al.~\cite{czolbe2020loss} introduce a frequency-based reconstruction loss for variational autoencoders (VAEs) using discrete Fourier transformation. Similarly, Cai et al. \cite{cai2021frequency} improve identity-preserving image generation by constraining their framework in pixel and Fourier spectral spaces. Nonetheless, these methods are not suitable for NST as perceptual losses are in the feature latent space, not input or output dimensions. We propose contrastive learning for aesthetic features, operating directly in the latent space and significantly reducing computational costs while extracting delicate aesthetic features.


\section{Method} \label{sec:Method}
Here, we introduce a novel methodology that substantially enhances the quality of synthesized images by effectively leveraging the potential of OctConv and decomposing feature maps according to their respective frequencies. The following sections provide a comprehensive examination of the proposed approach and its underlying principles.

\begin{figure}[t!]
    \centering
    \includegraphics[width=0.48\textwidth]{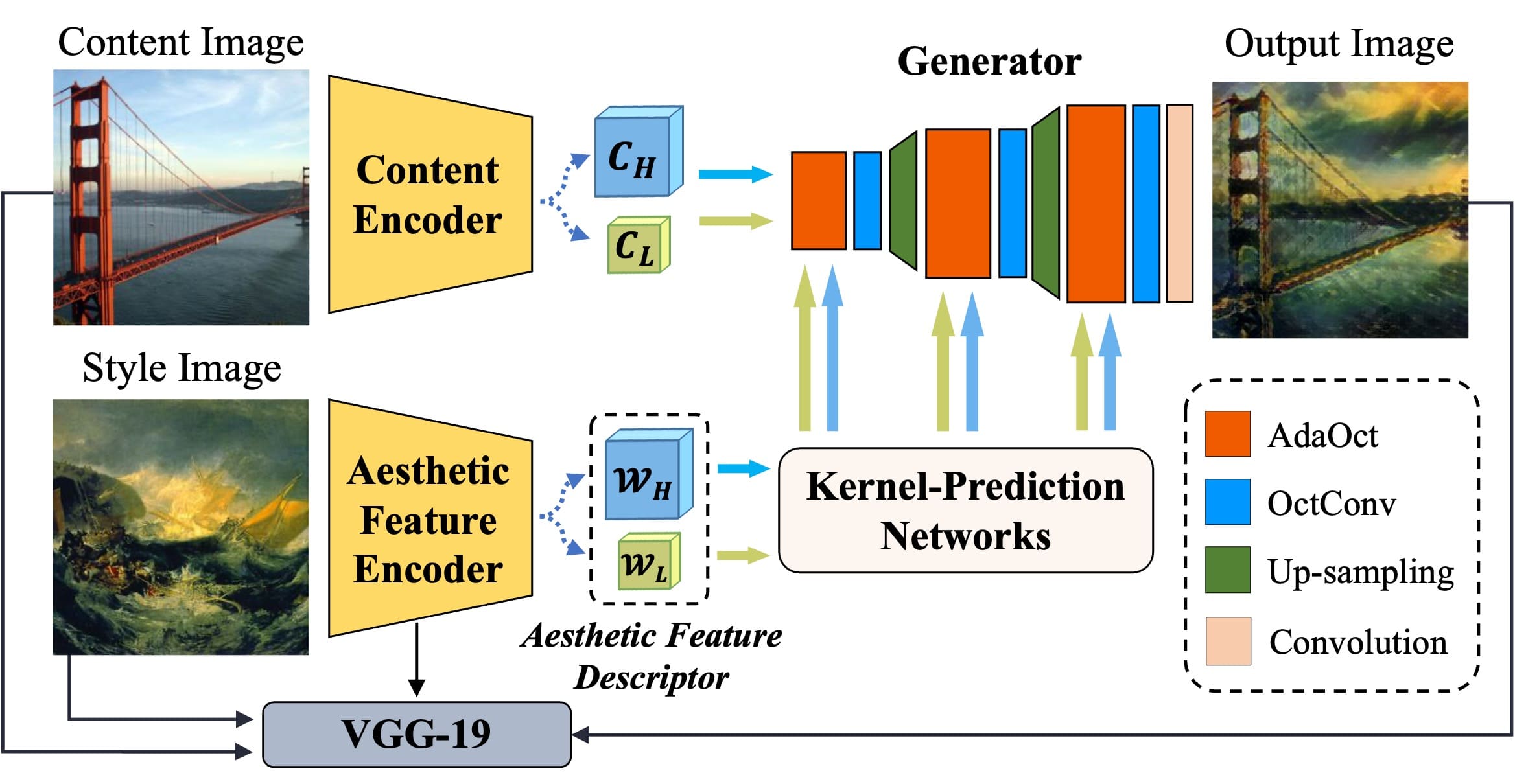}
    \caption{The entire AesFA architecture for aesthetic feature-aware NST. The blue and green arrows indicate the high- and low-frequency feature processes, respectively.}
    \label{fig:fig2}
\end{figure}

\subsection{Architecture Overview}
As depicted in Figure~\ref{fig:fig2}, the AesFA architecture comprises three primary components: a content encoder $E_c$, an aesthetic feature encoder $E_{aes}$ in conjunction with kernel-prediction networks $\mathcal{K}$, and a generator $G$. Specifically, the input contents are encoded via a content encoder $E_c$ and subsequently decomposed into two feature maps containing distinct frequency components. Meanwhile, the style images are processed through the aesthetic feature encoder $E_{aes}$ to encapsulate higher-level aesthetic feature information. Employing the aesthetic feature descriptor $\mathcal{W}$, which is encoded by the aesthetic feature encoder, the kernel-prediction networks $\mathcal{K}$ predict \textit{aesthetic feature-aware convolutional kernels and biases} for each respective spatial resolution. These predictions then are integrated into the generator alongside the decomposed content latent features. Within the generator, the content features merge with the predicted \textit{aesthetic feature-aware convolutional kernels and biases} for each corresponding frequency using AdaOct. In the terminal layer, the synthesized high- and low-frequency images are amalgamated to produce a single style-transferred output. To summarize, the overarching pipeline proceeds as follows:
\begin{enumerate}
    \item Encode two decomposed features $\mathcal{C}_{H}$, $\mathcal{C}_{L}$ (both the high frequency and low frequency) from the content image $\mathit{C}$ using the content encoder. To encode the aesthetic feature descriptor $\mathcal{W}$, the style image $\mathit{S}$ is fed to the aesthetic feature encoder $\mathit{E_{aes}}$. 
\begin{equation}
    \small
    \mathcal{C}_{H},\mathcal{C}_{L} := E_c(C), \,\,\,\,\,
    \mathcal{W}_{H},\mathcal{W}_{L} := E_{aes}(S)
\end{equation}
    \item Predict the \textit{aesthetic feature-aware kernels and biases} from the given aesthetic style descriptor $\mathcal{W}$ using kernel-prediction networks $\mathcal{K}$. These kernels and biases will be used in the n-$\mathit{th}$ layer of the generator.
\begin{equation}
    \small
    k_{n,H},b_{n,H} := K_{n, H}(\mathcal{W}_H), \,\,\,k_{n,L},b_{n,L} := K_{n, L}(\mathcal{W}_L)
\end{equation}
    \item Infuse aesthetic styles with contents in the generator $\mathit{G}$, creating style-transferred output $\mathit{O}$.
\begin{equation}
    \small
    O := G(\mathcal{C}_{H}, \mathcal{C}_{L}, \{k_{n,H}, b_{n,H}\}, \{k_{n,L}, b_{n,L}\})
\end{equation}
\end{enumerate}

\subsection{Frequency Decomposition Networks}\label{sec:encoder}
\textbf{Octave Convolution.} A pivotal aspect of the OctConv operator is its capacity to factorize mixed feature maps by their frequencies while concurrently facilitating efficient communication between high- and low-frequency components. Low-frequency feature maps in OctConv have their spatial resolution reduced by one octave~\cite{lindeberg2013scale}, where an octave is a spatial dimension divided by a power of two. In this study, a value of 2 was chosen for simplicity. Given input and output of OctConv as $\mathit{X = \{X_H,X_L\}}$ and $\mathit{Y=\{Y_H,Y_L\}}$, the forward pass is defined as:
\begin{equation} \begin{gathered}
    \small
    \mathit{Y}_{H} = \mathit{f}(\mathit{X}_H ; \mathit{W}_{H\rightarrow H}) + \mathit{f}(upsample(\mathit{X}_L, 2) ; \mathit{W}_{L \rightarrow H}) \\
    \mathit{Y}_{L} = \mathit{f}(\mathit{X}_L ; \mathit{W}_{L\rightarrow L}) + \mathit{f}(pool(\mathit{X}_H, 2) ; \mathit{W}_{H \rightarrow L}),
\end{gathered}
\end{equation}

\noindent where $\small \mathit{f}(\mathit{X};\mathit{W})$ represents a convolution with parameters $\mathit{W}$. Then, $\textit{pool(X, 2)}$ and $\textit{upsample(X, 2)}$ denote an average pooling operation with kernel size 2$\times$2 with a stride of 2 and an upsampling operation by 2 using the nearest interpolation, respectively. Empirical findings indicate that employing OctConv with half the channels for each frequency ($\alpha=0.5$) yields optimal performance.

\begin{figure}[!t]
\centering
\includegraphics[width=0.49\textwidth]{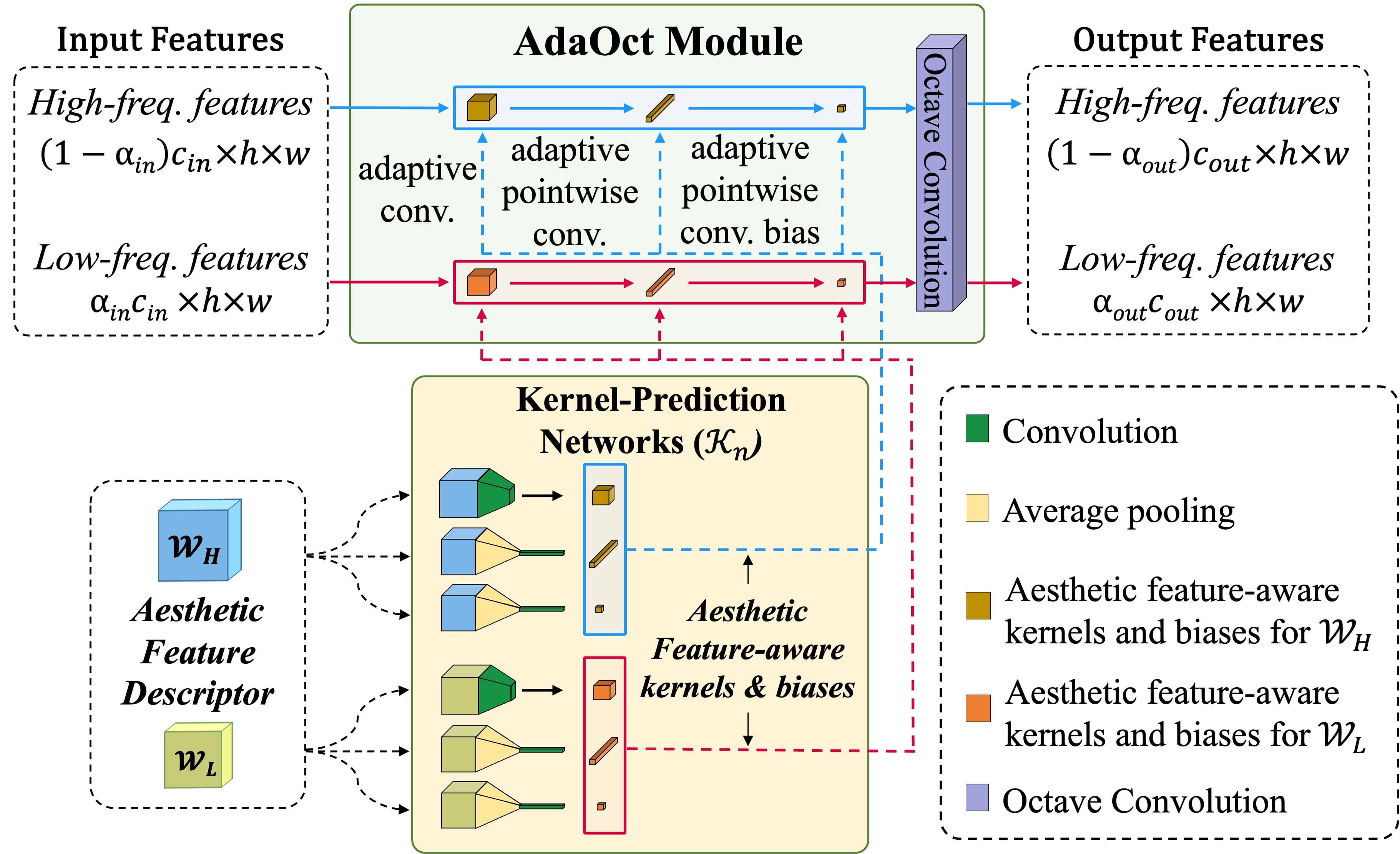}
\caption{The detailed design of the Adaptive Octave Convolutions (AdaOct) used in AesFA.}
\label{fig:fig3}
\end{figure}

\textbf{Content and Aesthetic Feature Encoders.} Both proposed encoders improve upon MobileNet~\cite{howard2017mobilenets} by replacing all convolutions with OctConv to factorize feature maps by their frequency, thereby reducing network redundancy while maintaining simplicity and efficiency. Spatial reduction in the low-frequency branch expands the receptive field, capturing more contextual information from distant locations and improving performance with greater image resolution. In contrast to the original OctConv, the upsampling order is adjusted to address checkerboard artifacts.

\subsection{Aesthetic Feature-Aware Stylization}\label{sec:generator}

\textbf{Kernel-Prediction Networks.} To effectively apply the aesthetic feature descriptor to content features, we present an approach using kernel-prediction networks similar to those of AdaConv. These networks predict \textit{aesthetic feature-aware kernels and biases} in a depthwise-separable manner, corresponding to frequency and spatial resolution. The \textit{aesthetic feature-aware kernels and biases} comprise depthwise convolution components, pointwise convolution components, and per-channel biases. This approach diverges from the original kernel-prediction network utilized in AdaConv by predicting \textit{aesthetic feature-aware kernels and biases} from both high- and low-frequency aesthetic feature descriptors.

\textbf{Adaptive Octave Convolutions.} To efficiently integrate frequency-decomposed contents with the predicted \textit{aesthetic feature-aware kernels and biases}, we begin with AdaConv's original architecture. However, instead of using it directly, we employ AdaOct followed by an OctConv rather than the standard convolutions outlined in AdaConv. The active interactions between two frequencies that occur in OctConv could further enhance aesthetic stylization quality while reducing the total computational redundancy and unwanted artifacts. Figure~\ref{fig:fig3} provides a detailed overview of our AdaOct module.

The generator comprises three layers, each consisting of an AdaOct module and a standard OctConv block, followed by an upsampling operator. The role of the standard OctConv after the AdaOct module is to learn style-independent kernels, which aid in the reconstruction of high-quality images. When convolving with \textit{aesthetic feature-aware kernels and biases}, the input channels are grouped into $\mathit{n_g}$ independent clusters. The network then applies separate spatial and pointwise kernels to learn aesthetic features, such as the microstructure of the texture, as well as cross-channel correlations within the input features. The value of $n_g$ remains consistent for all \textit{aesthetic feature-aware kernels and biases}, while the remaining parameters adhere to those defined by AdaConv. Notably, AdaConv requires fixed dimensions for style images due to its fully connected layer, while, being fully convolutional, AesFA handles inputs of varying dimensions for both content and style images.


\begin{figure}[!t]
\centering
\includegraphics[width=0.49\textwidth, height=3.85cm]{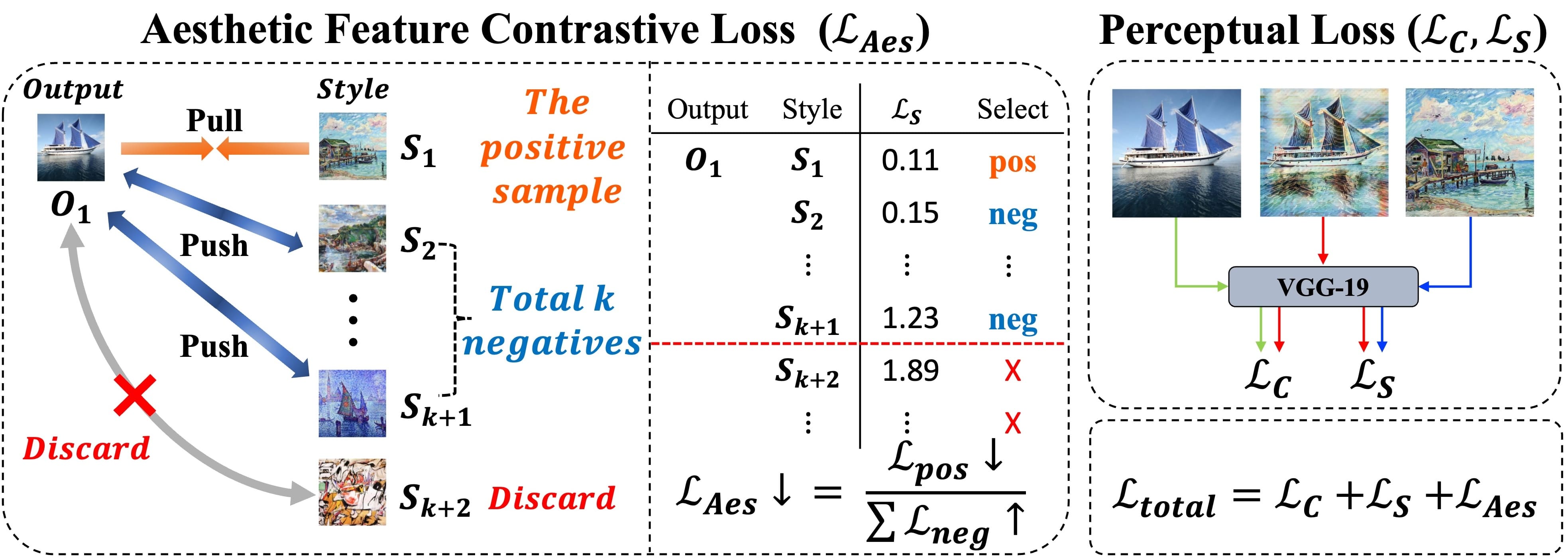}
\caption{
Illustration of the \textit{aesthetic style contrastive loss} in a toy example alongside the other training losses employed in AesFA.
}
\label{fig:fig4}
\end{figure}

\subsection{Aesthetic Feature Contrastive Learning}
A singular perceptual loss is insufficient for extracting and expressing intricate aesthetic-style representations (refer to Figure~\ref{fig:fig8}). To address this limitation, we adopt and improve the contrastive learning approach from MicroAST~\cite{wang2023microast}. Our enhanced loss, termed \textit{Aesthetic Feature Contrastive Loss} ($\mathcal{L}_{Aes}$), follows the contrastive learning principle of maintaining proximity between data and their corresponding ``positive'' samples while distancing them from other instances deemed as ``negatives'' in the representation space. Consequently, the selection of ``positive'' and ``negative'' samples is pivotal for the success of contrastive learning.

Despite its remarkable progress, MicroAST calculates contrastive loss using all negative samples in a mini-batch, making it computationally expensive, particularly at higher resolutions. Intuitively, the nearest sample from the positive offers the most distinctive information. Inspired by hard negative mining techniques~\cite{robinson2020contrastive}, we redefined ``negative'' samples as a subset of the entire negative sample pool, comprising the $k$-th nearest negative samples to the style-transferred output image. For each style-transferred output, the corresponding style image is designated as its positive sample. Meanwhile, the remaining outputs are treated as its \textit{pseudo}-samples. Perceptual losses are defined as the distance between positive and \textit{pseudo}-negative samples, which is calculated by the Exact Feature Distribution Matching (EFDM) algorithm \cite{zhang2022exact} using the pre-trained VGG-19 network. 

The \textit{pseudo}-style perceptual losses are arranged in ascending order, and the top $k$ \textit{pseudo}-negative samples are selected to compute the final aesthetic feature contrastive loss. The variable $k$ represents a design choice and can be arbitrarily large, subject to the mini-batch size. In this study, we found that using the nearest style image (i.e., $k=1$) yields the best performance. The aesthetic feature contrastive loss is computed at each layer of the entire encoder and on both the high- and low-frequency branches. The formal aesthetic feature contrastive loss is formalized as follows:

\begin{equation}\label{eq:contrastive}
\begin{gathered} \small
\mathcal{L}_{Aes} := \sum_{l=1}^{}\mathcal{L}_{Aes, l, High} + \sum_{l=1}^{}\mathcal{L}_{Aes, l, Low} \\
\mathcal{L}_{Aes, l} := \\ \small\sum_{i=1}^{N}\frac{\small ||\mathcal{F}_l(O_i)-\mathrm{EFDM}(\mathcal{F}_l(O_i), \mathcal{F}_l(S_{pos, i})) ||_2}{\small \sum_{j=1}^{k}||\mathcal{F}_l(O_i)-\mathrm{EFDM}(\mathcal{F}_l(O_i), \mathcal{F}_l(S_{neg, j}))||_2},
\end{gathered}
\end{equation}

\noindent where $\mathcal{F}_l(x)$ represents the feature activations of $\mathit{l}$-th layer in our encoder given the input $\mathit{x}$ and $N$ mini-batch. ${S_{pos}}$ and ${S_{neg}}$ represent the positive and negative samples for each style-transferred output $O$, respectively.

\subsection{Training Losses}
\textbf{Perceptual Loss.} In accordance with previous studies \cite{gatys2016image, johnson2016perceptual}, we use the pre-trained VGG-19 model to compute the perceptual loss, which consists of both the content and style losses. However, this work redefines the style loss $\mathcal{L}_S$ as in the case of EFDM. Given $I$ representing the stylized output and $y$ denoting the reference image, the final perceptual losses are as follows:

\begin{equation}
\begin{gathered}
\small \mathcal{L}_C = ||f_3(I)-f_3(y)||_2,\\
\small \mathcal{L}_S = \sum_{n=1}^{4}||f_n(I)- \mathrm{EFDM}(f_n(I), f_n(y))||_2,
\end{gathered}
\end{equation}

\noindent where $\mathit{f_n}$ symbolizes the $\mathit{n}$-th layer in the VGG-19 model. The content loss is computed at the \{\textit{conv}3\_1\} layer in VGG-19, while the style loss is calculated at the \{\textit{conv1\_1, conv2\_1, conv3\_1, conv}4\_1\}. It is important to note the VGG-19 model is used solely during training and is entirely excluded from the inference process.

\textbf{Total Loss.} Considering all of the aforementioned losses, the total loss is formalized as:
\begin{equation}
    \mathcal{L}_{total}=\lambda_C \mathcal{L}_C+ \lambda_S \mathcal{L}_S+\lambda_{Aes} \mathcal{L}_{Aes},
\end{equation}
\noindent where $\mathit{\lambda_C, \lambda_S}$,  and $\mathit{\lambda_{Aes}}$ are the weighting hyperparameters for each loss. In this paper, we use $\mathit{\lambda_C= }$1 , ${\lambda_S= }$10,  and $\mathit{\lambda_{Aes}}=$5. Figure~\ref{fig:fig8} describes the impact of each hyperparameter.

\subsection{Implementation Details} \label{sec:implementation}
To train our model, we use the COCO dataset~\cite{lin2014microsoft} as content images and the WikiArt dataset \cite{phillips2011wiki} as style images. During training, images are rescaled to 512 pixels while maintaining the original aspect ratio then randomly cropped to 256$\times$256 pixels for augmentation. The model is trained using the Adam optimizer \cite{kingma2014adam} with a learning rate of 0.0001 and a batch size of 8 for 160,000 iterations. The aesthetic feature has dimensions of (256, 3, 3) for both high- and low-frequency components. All experiments were conducted using the PyTorch framework \cite{paszke2019pytorch} on a single NVIDIA A100(40G) GPU.

\begin{figure*}[!t]
    \centering
    \includegraphics[width=\textwidth, height=4.5cm]{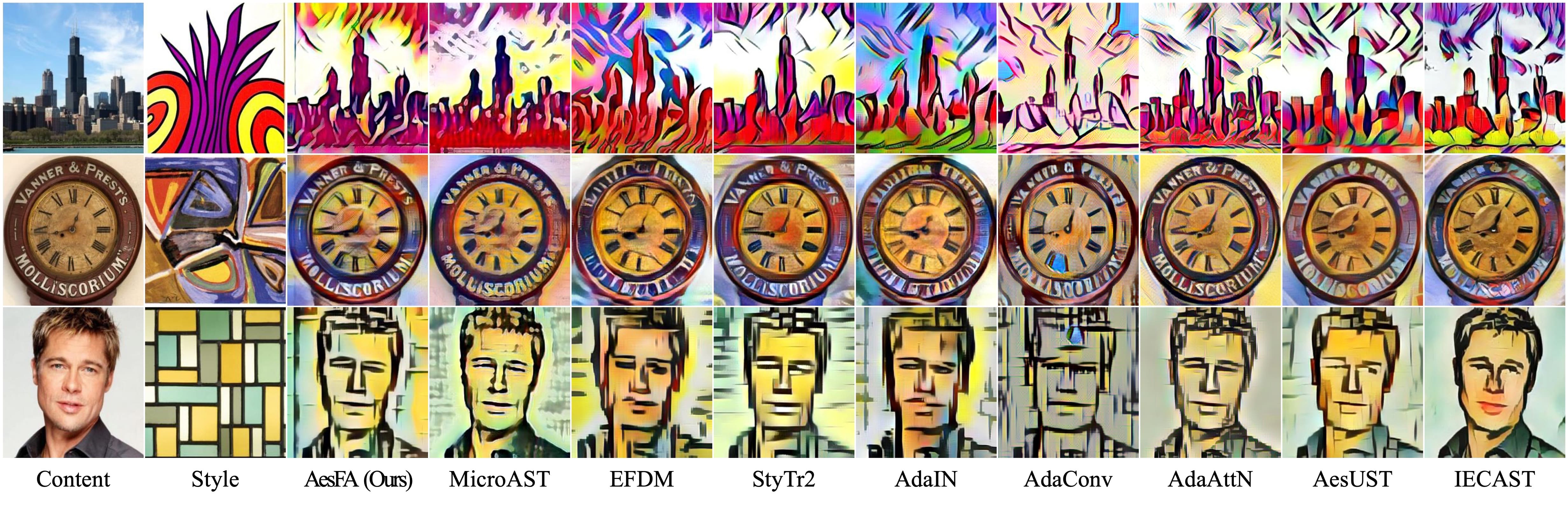} 
    \caption{Qualitative comparison with various NST algorithms in 256 pixel resolution. Each column shows the stylized images of different state-of-the-art models.}
    \label{fig:fig5}
\end{figure*}

\begin{figure*}[!t]
    \centering
    \includegraphics[width=\textwidth, height=5.7cm]{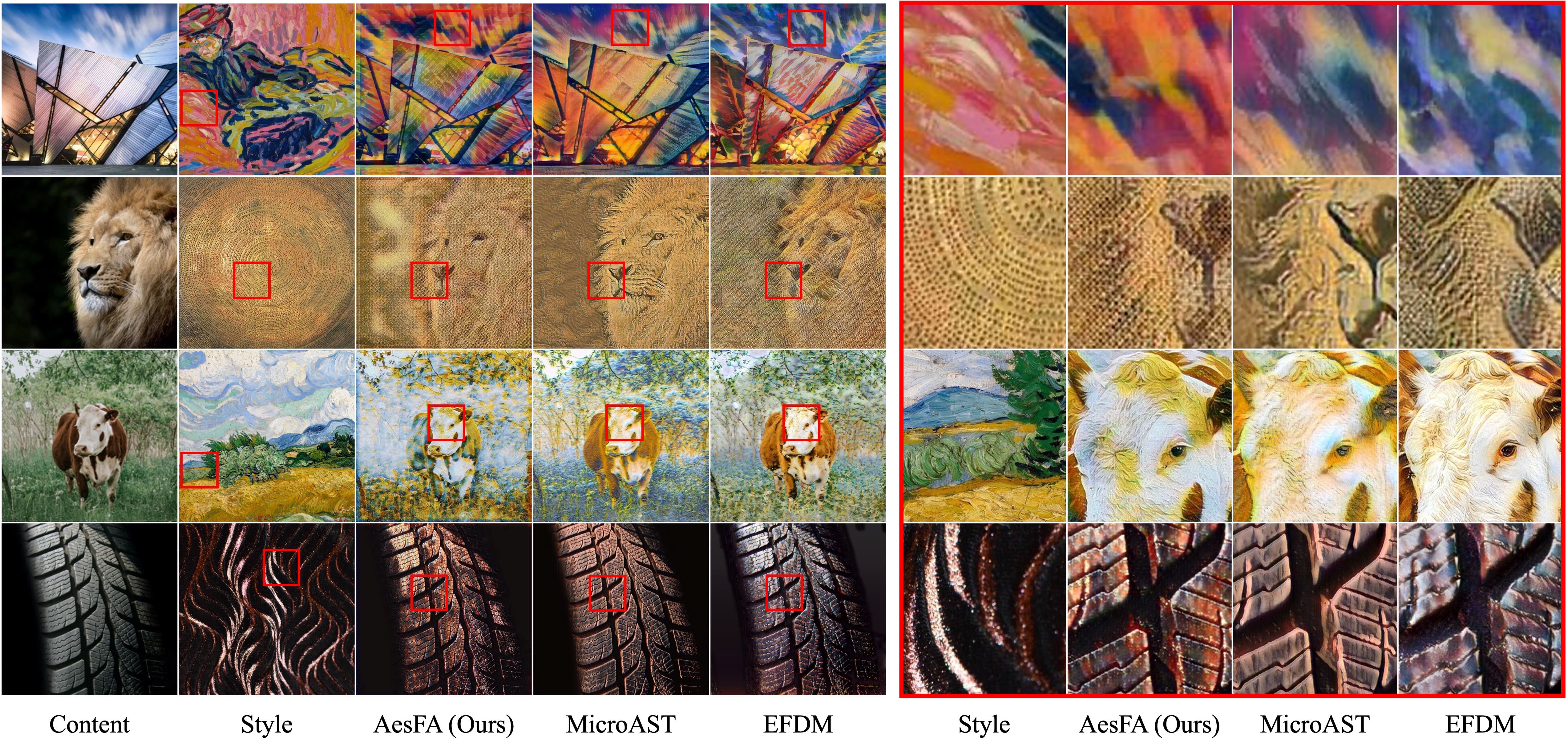} 
    \caption{Qualitative comparison with various NST algorithms in 512 (first and second row) and 2K (2048$\times$2048; third and fourth row) resolution. First row: tones, second row: texture, third row: brushstrokes, and fourth row: grains. In all aesthetic features, our AesFA method outperforms.}
    \label{fig:fig6}
\end{figure*}

\section{Experimental Results}
In this section, the proposed model's validity is assessed both qualitatively and quantitatively in comparison to state-of-the-art NST approaches, including AesUST~\cite{wang2022aesust}, AdaIN~\cite{huang2017arbitrary}, AdaConv~\cite{chandran2021adaptive}, MicroAST~\cite{wang2023microast}, EFDM~\cite{zhang2022exact}, AdaAttn~\cite{liu2021adaattn}, IECAST~\cite{chen2021artistic}, and StyTr$^2$~\cite{deng2021stytr2}. We conduct experiments on a range of image resolutions, spanning from small resolutions of 256 pixels to ultra-high 4K resolution. A total of 10 content images and 20 style images are randomly selected for the tests, including images sourced from WikiArt and \textit{pexels.com} for ultra-high resolution images. For each spatial resolution, we generate 200 test results. The results for these methods are acquired by retraining the respective author-released codes using default configurations. 

\begin{figure*}[!t]
    \centering
    \includegraphics[width=\textwidth, height=7cm]{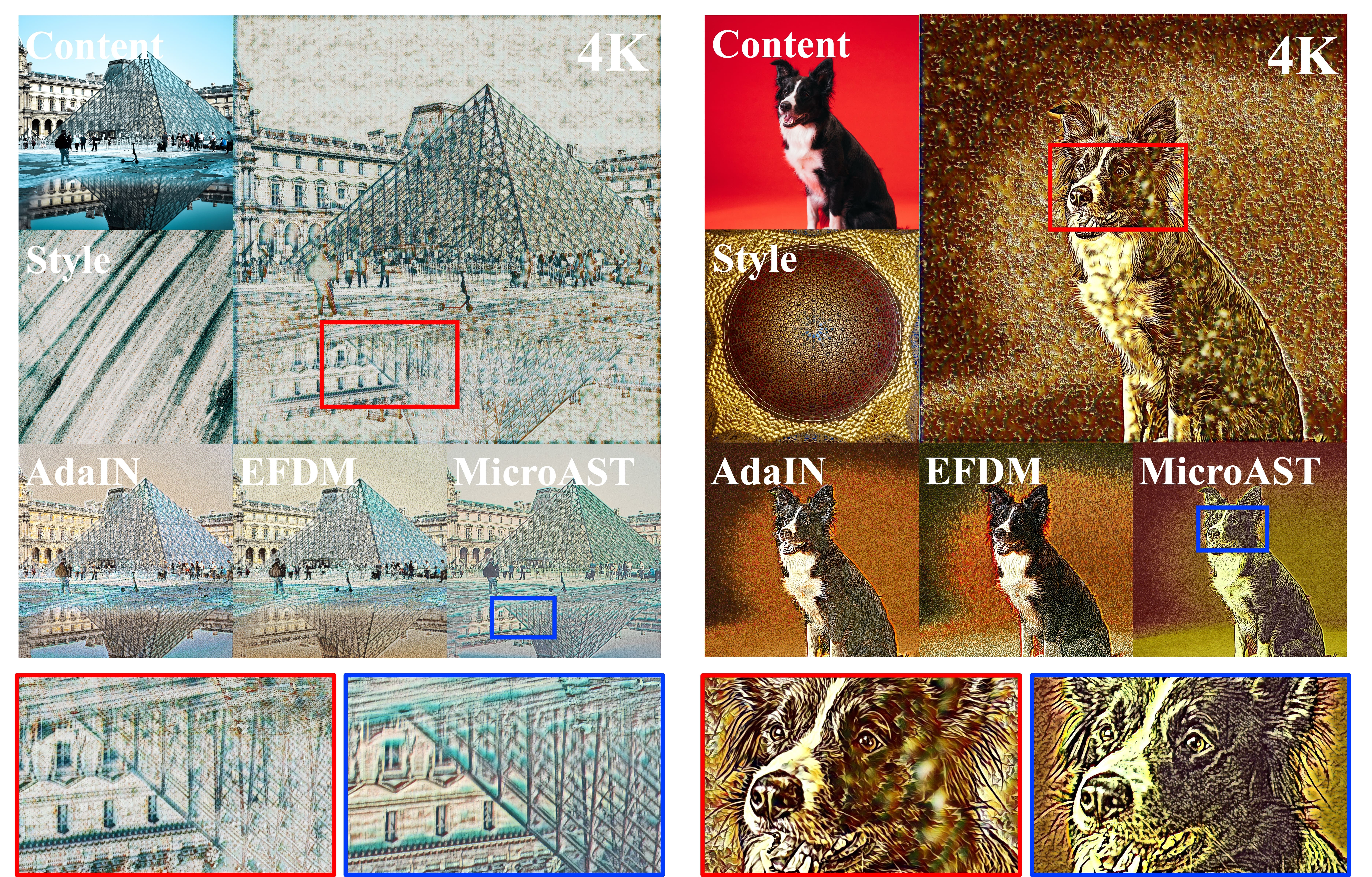} 
    \caption{Ultra-high resolution (4K; 4096$\times$4096) comparison. The top-left images are content and style images, and the top right displays our synthesized image. The bottom-left image is the magnified image of our result, and the image on the right is the magnified result from the current state-of-the-art model. Our model outperforms in terms of aesthetic features (e.g., brushstrokes; texture; tones). Zoom in for details.} 
    \label{fig:fig7}
\end{figure*}

\textbf{Qualitative Comparisons.} As described in Figure~\ref{fig:fig5}, AesFA qualitatively outperforms eight state-of-the-art NST techniques in terms of aesthetics while maintaining the essential content semantics. AesFA excels in the transfer of unique local aesthetic structural elements from the style image to the content image at all spatial resolutions. Notably, Figure~\ref{fig:fig6} demonstrates that our method can faithfully show aesthetic feature-aware style transfer in terms of tones (first row), texture (second row), brushstrokes (third row), and grains (fourth row). Figure~\ref{fig:fig7} also shows that AesFA excels in transferring the local structure of the style image to the content image in ultra-high resolution (e.g., 4K). MicroAST, for example, suffers from poor aesthetic stylizations and low image quality (blue box in Figure~\ref{fig:fig7}). In contrast, AesFA achieves promising outputs with higher image quality (red box in Figure~\ref{fig:fig7}). Additional results are demonstrated in the supplementary materials.

\begin{table}[!ht]
\renewcommand*{\arraystretch}{1.2}
\setlength{\tabcolsep}{5pt}
\centering \resizebox{0.48\textwidth}{!}{
\huge
\begin{tabular}{c|c|ccccc}
\toprule
Resolution & Method & Style Loss ($\downarrow$) & LPIPS ($\downarrow$) & SSIM ($\uparrow$) & Time (sec, $\downarrow$)  & Pref. (\%, $\uparrow$) \\ \hline \hline
$256^2$        & AdaConv (\citeyear{chandran2021adaptive})      & 0.936          & 0.379 & 0.246 & 0.493 & 2.50           \\
           & AdaIN (\citeyear{huang2017arbitrary})       & 0.727          & 0.371 & 0.230 & 0.011 & 11.40           \\
           & MicroAST (\citeyear{wang2023microast})     & 1.189          & 0.372 & 0.408 & \textbf{0.007} & 12.35           \\
           & EFDM (\citeyear{zhang2022exact})        & 0.720          & 0.378 & 0.212 & 0.011 & 6.00           \\
           & AdaAttn (\citeyear{liu2021adaattn})     & 0.993          & 0.390 & \textbf{0.468} & 0.027 & 8.23           \\
           & Aes-UST (\citeyear{wang2022aesust})     & 0.731          & 0.372 & 0.355 & 0.019 & 9.18            \\
           & IECAST (\citeyear{chen2021artistic})      & 0.984          & 0.392 & 0.342 & 0.025 & 13.28           \\
           & StyTr$^2$ (\citeyear{deng2021stytr2})   & \textbf{0.581} & 0.377 & 0.450 & 0.038 & 7.90          \\ \cline{2-7}
           & \textbf{AesFA (Ours)} & 0.692          & \textbf{0.368} & 0.417 & 0.016 & \textbf{32.90}           \\ \hline
$1024^2$       & AdaConv (\citeyear{chandran2021adaptive})     & N/A            & N/A   & N/A   & N/A   &      \textemdash      \\
           & AdaIN (\citeyear{huang2017arbitrary})        & 0.373          & 0.399 & 0.336 & 0.014 & 4.73           \\
           & MicroAST (\citeyear{wang2023microast})    & 0.531          & 0.400 & 0.430 & \textbf{0.011} & 15.50  \\
           & EFDM (\citeyear{zhang2022exact})        & 0.342          & 0.401 & 0.313 & 0.013 & 6.33           \\
           & AdaAttn (\citeyear{liu2021adaattn})     & 0.596          & 0.459 & 0.484 & 0.060 & 14.10           \\
           & Aes-UST (\citeyear{wang2022aesust})     & 0.423          & 0.420 & 0.455 & 0.024 & 14.58           \\
           & IECAST (\citeyear{chen2021artistic})      & 0.554          & 0.438 & 0.438 & 0.015 & 12.38           \\
           & StyTr$^2$ (\citeyear{deng2021stytr2})   & 0.288          & 0.411 & \textbf{0.475} & 1.241 & 8.88           \\ \cline{2-7}
           & \textbf{AesFA (Ours)} & \textbf{0.283} & \textbf{0.392} & 0.405 & 0.020 & \textbf{25.03}           \\ \hline
$2048^2 (2K)$   & AdaConv (\citeyear{chandran2021adaptive})      & N/A            & N/A   & N/A   & N/A   &      \textemdash      \\
           & AdaIN (\citeyear{huang2017arbitrary})       & 0.531          & 0.443 & 0.311 & \textbf{0.013} & 19.00     \\
           & MicroAST (\citeyear{wang2023microast})    & 0.709          & 0.447 & 0.406 & 0.014 &  15.18          \\
           & EFDM (\citeyear{zhang2022exact})       & 0.475          & 0.448 & 0.299 & 0.018 & 14.90           \\
           & AdaAttn (\citeyear{liu2021adaattn})     & OOM            & OOM   & OOM   & OOM   &       \textemdash     \\
           & Aes-UST (\citeyear{wang2022aesust})     & 0.754          & 0.458 & \textbf{0.441} & 0.028 & 16.50           \\
           & IECAST (\citeyear{chen2021artistic})       & OOM            & OOM   & OOM   & OOM   &      \textemdash      \\
           & StyTr$^2$ (\citeyear{deng2021stytr2})   & OOM            & OOM   & OOM   & OOM   &      \textemdash      \\ \cline{2-7}
           & \textbf{AesFA (Ours)} & \textbf{0.404} & \textbf{0.435} & 0.378 & 0.020 &  \textbf{34.48}          \\ \hline
$4096^2 (4K)$   & AdaConv (\citeyear{chandran2021adaptive})      & N/A            & N/A   & N/A   & N/A   &      \textemdash      \\
           & AdaIN (\citeyear{huang2017arbitrary})      & 0.428          & 0.376 & 0.384 & 0.022 & 15.53           \\
           & MicroAST (\citeyear{wang2023microast})    & 0.453          & \textbf{0.371} & \textbf{0.477} & \textbf{0.019} & 14.88           \\
           & EFDM ~(\citeyear{zhang2022exact})        & 0.412          & 0.379 & 0.382 & 0.028 & 19.00           \\
           & AdaAttn (\citeyear{liu2021adaattn})     & OOM            & OOM   & OOM   & OOM   &       \textemdash     \\
           & Aes-UST (~\citeyear{wang2022aesust})     & OOM            & OOM   & OOM   & OOM   &      \textemdash      \\
           & IECAST (\citeyear{chen2021artistic})       & OOM            & OOM   & OOM   & OOM   &      \textemdash      \\
           & StyTr$^2$ (\citeyear{deng2021stytr2})       & OOM            & OOM   & OOM   & OOM   &     \textemdash       \\ \cline{2-7}
           & \textbf{AesFA (Ours)} & \textbf{0.216} & 0.373 & 0.469 & \textbf{0.020} & \textbf{50.60}           \\ \toprule
\end{tabular}}
\caption{Quantitative comparison with various state-of-the-art NST algorithms. ``N/A'' means ``Not applicable at this resolution'' and ``OOM'' stands for ``Out of GPU memory''.} 
\label{tab:quantitative}
\end{table}

\textbf{Quantitative Comparisons.} To ensure a comprehensive and effective quantitative comparison, we employ three evaluation metrics: the average SSIM~\cite{wang2004image}, the style perceptual loss measured in VGG space~\cite{johnson2016perceptual}, and the Learned Perceptual Image Patch Similarity (LPIPS)~\cite{zhang2018unreasonable}. These metrics are used to evaluate the stylization quality in terms of its ability to preserve content and achieve desirable stylization effects. Table~\ref{tab:quantitative} shows the quantitative results with various state-of-the-art NST models. Compared to the other techniques, AesFA accomplishes the highest or at least comparable score along all evaluation metrics regardless of image spatial resolution, rendering a single image in less than 0.02 seconds.

\textbf{User Study.} Evaluating the outcomes of stylization is a highly subjective matter. Hence, we have conducted a user study for the nine approaches. We randomly show each participant 20 ballots (4 ballots for each resolution) containing the content, style, and nine outputs. For each ballot, participants were given unlimited time to select their favorite output in terms of aesthetically pleasing stylization and content preservation. We collected 1,580 valid votes from 79 subjects. The preference percentage of each method for each resolution is included in the last column of Table~\ref{tab:quantitative}. The user study results demonstrate that our stylized images are more appealing than or at least comparable to the competitors.

\subsection{Ablation Studies}\label{sec:ablations}

\begin{figure}[t!]
    \centering
    \includegraphics[width=0.48\textwidth, height=5cm]{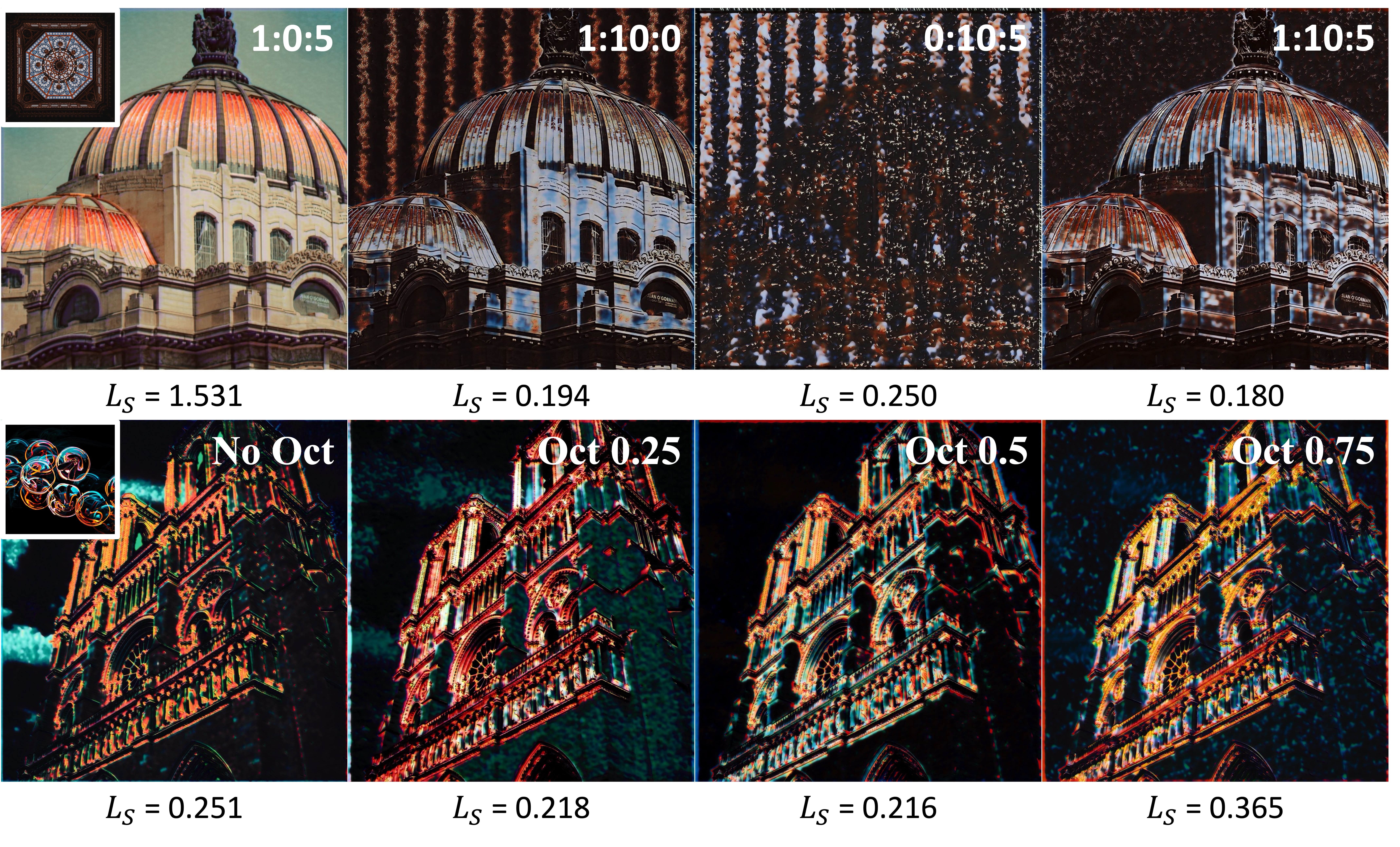}
    \caption{Top: The effectiveness of the proposed \textit{aesthetic feature contrastive loss}. Bottom: The effectiveness of the $\alpha$ value of OctConv. $\mathcal{L}_{S}$ denotes the style perceptual loss.} 
    \label{fig:fig8}
\end{figure}

We also have conducted a series of ablation studies to provide justification for the architectural decisions employed and to highlight their effectiveness. We first explore the effect of \textit{aesthetic feature contrastive loss,} $\mathcal{L}_{Aes}$ in Figure~\ref{fig:fig8}. When training without $\mathcal{L}_{Aes}$, the stylization quality of the proposed model drastically degrades, and unsatisfactory artifacts appear (e.g., the stripe pattern in the background). This shows that the newly devised loss revealed by AesFA plays an important role in expressing aesthetic features and eliminating artifacts. Notably, we changed the alpha value ($\alpha$) of the OctConv, which denotes the ratio of the number of low-frequency channels to the total-frequency channels. Results show that our model with $\alpha=0.5$ performs the best qualitatively (Figure~\ref{fig:fig8}) and quantitatively, significantly reducing artifacts in the background and enhancing stylization quality. Meanwhile, the model with standard convolutions (No Oct) shows undesirable artifacts in the background and poor quality of colorization. Detailed descriptions and images of high- and low-frequency component images for each setting are provided in the supplementary materials. 

\begin{figure}[t!]
\centerline{\includegraphics[width=0.48\textwidth, height=5cm]{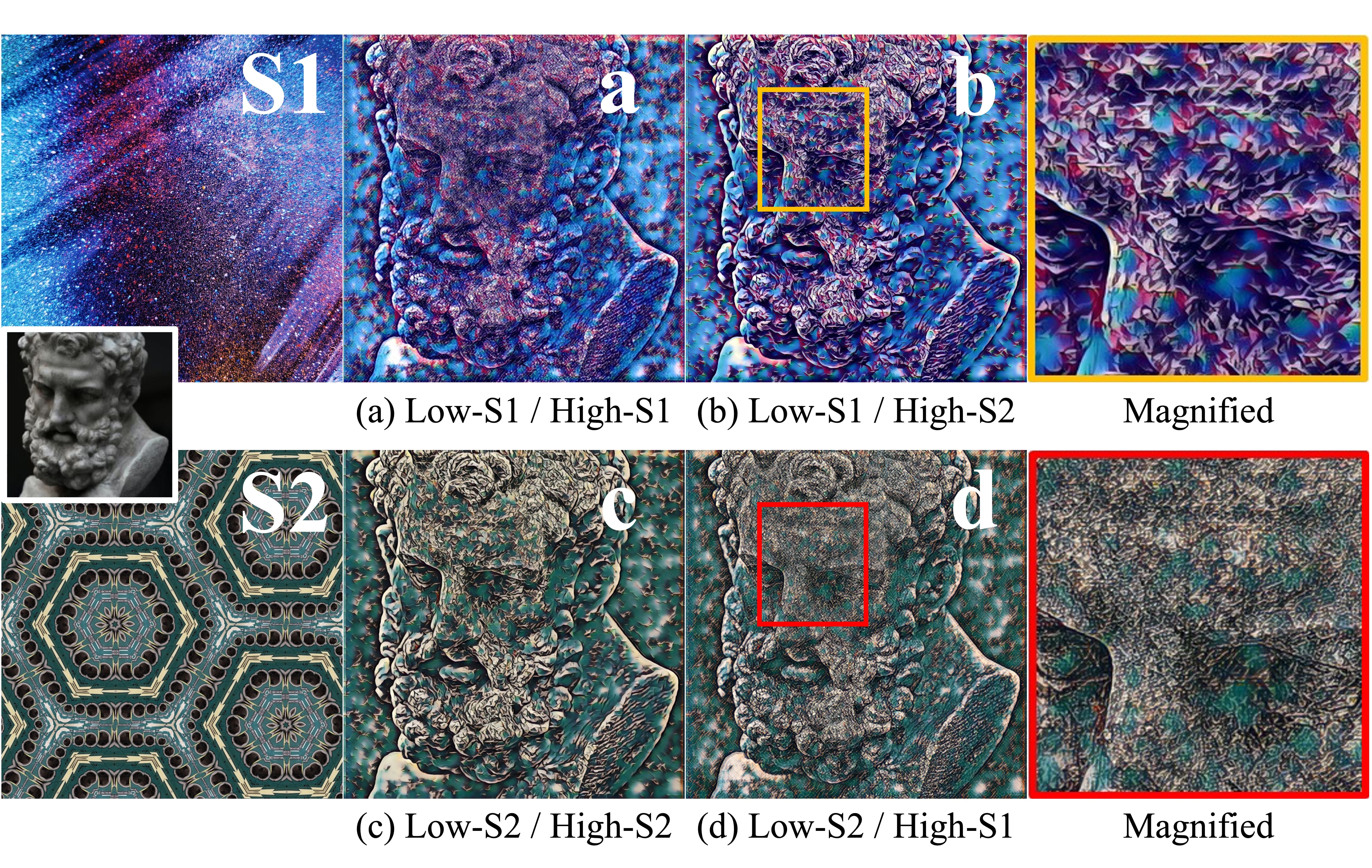}}
    \caption{The style blending results generated by AesFA with different style images. Zoom in for details.}
    \label{fig:fig9}
\end{figure}

Figure~\ref{fig:fig9} shows the style blending, i.e., using the low- and high-frequency style information from different style images. Sub-figures (a)-(d) show different combinations of origins for low- and high-frequency style information. For instance, ``(b) Low-S1 / High-S2" indicates that we use the low-frequency style information from image ``S1" and the high-frequency style information from image ``S2".


\section{Conclusion}
In this study, we propose AesFA, a lightweight and effective model for aesthetic feature-aware NST. Unlike existing models, AesFA decomposes the image by frequencies and infuses it with corresponding aesthetic features. We introduce a new aesthetic feature contrastive loss by leveraging pretrained VGGs to guide stylization effectively. Our experiments demonstrate that the model and new loss significantly enhance the quality of generated images \textit{regardless of resolution}. Furthermore, AesFA achieves stylized output in less than 0.02 seconds, making it suitable for real-time ultra-high resolution rendering (4K) applications.

\section*{Acknowledgements} This work was supported by the U.S. Department of Energy (DOE), Office of Science (SC), Advanced Scientific Computing Research program under award DE-SC-0012704 and used resources of the National Energy Research Scientific Computing Center, a DOE Office of Science User Facility supported by the Office of Science of the U.S. Department of Energy under Contract No. DE-AC02-05CH11231 using NERSC award ASCR-ERCAP0023081.

This work was also supported by the National Research Foundation of Korea (NRF) grant funded by the Korea government (MSIT) (No. 2021R1C1C1006503, 2021K1A3A1A2103751212, 2021M3E5D2A01022515, RS-2023-00266787, RS-2023-00265406), by Creative-Pioneering Researchers Program through Seoul National University(No. 200-20230058), by Semi-Supervised Learning Research Grant by SAMSUNG(No.A0426-20220118), and by Institute of Information \& communications Technology Planning \& Evaluation (IITP) grant funded by the Korea government (MSIT) [NO.2021-0-01343, Artificial Intelligence Graduate School Program (Seoul National University)].

\bibliography{aaai24}

\begin{thebibliography}{60}
\providecommand{\natexlab}[1]{#1}

\bibitem[{Akbari et~al.(2020)Akbari, Liang, Han, and Tu}]{akbari2020generalized}
Akbari, M.; Liang, J.; Han, J.; and Tu, C. 2020.
\newblock Generalized octave convolutions for learned multi-frequency image compression.
\newblock \emph{arXiv preprint arXiv:2002.10032}.

\bibitem[{An et~al.(2023)An, Li, Huang, Ma, and Luo}]{an2023bigger}
An, J.; Li, T.; Huang, H.; Ma, J.; and Luo, J. 2023.
\newblock Is Bigger Always Better? An Empirical Study on Efficient Architectures for Style Transfer and Beyond.
\newblock In \emph{Proceedings of the IEEE/CVF Winter Conference on Applications of Computer Vision}, 4084--4094.

\bibitem[{Cai et~al.(2021)Cai, Zhang, Huang, Geng, Li, and Huang}]{cai2021frequency}
Cai, M.; Zhang, H.; Huang, H.; Geng, Q.; Li, Y.; and Huang, G. 2021.
\newblock Frequency domain image translation: More photo-realistic, better identity-preserving.
\newblock In \emph{Proceedings of the IEEE/CVF International Conference on Computer Vision}, 13930--13940.

\bibitem[{Chandran et~al.(2021)Chandran, Zoss, Gotardo, Gross, and Bradley}]{chandran2021adaptive}
Chandran, P.; Zoss, G.; Gotardo, P.; Gross, M.; and Bradley, D. 2021.
\newblock Adaptive convolutions for structure-aware style transfer.
\newblock In \emph{Proceedings of the IEEE/CVF conference on computer vision and pattern recognition}, 7972--7981.

\bibitem[{Chen et~al.(2018)Chen, Fan, Mallinar, Sercu, and Feris}]{chen2018big}
Chen, C.-F.; Fan, Q.; Mallinar, N.; Sercu, T.; and Feris, R. 2018.
\newblock Big-little net: An efficient multi-scale feature representation for visual and speech recognition.
\newblock \emph{arXiv preprint arXiv:1807.03848}.

\bibitem[{Chen et~al.(2021)Chen, Wang, Zhang, Zuo, Li, Xing, Lu et~al.}]{chen2021artistic}
Chen, H.; Wang, Z.; Zhang, H.; Zuo, Z.; Li, A.; Xing, W.; Lu, D.; et~al. 2021.
\newblock Artistic style transfer with internal-external learning and contrastive learning.
\newblock \emph{Advances in Neural Information Processing Systems}, 34: 26561--26573.

\bibitem[{Chen and Schmidt(2016)}]{chen2016fast}
Chen, T.~Q.; and Schmidt, M. 2016.
\newblock Fast patch-based style transfer of arbitrary style.
\newblock \emph{arXiv preprint arXiv:1612.04337}.

\bibitem[{Chen et~al.(2019)Chen, Fan, Xu, Yan, Kalantidis, Rohrbach, Yan, and Feng}]{chen2019drop}
Chen, Y.; Fan, H.; Xu, B.; Yan, Z.; Kalantidis, Y.; Rohrbach, M.; Yan, S.; and Feng, J. 2019.
\newblock Drop an octave: Reducing spatial redundancy in convolutional neural networks with octave convolution.
\newblock In \emph{Proceedings of the IEEE/CVF international conference on computer vision}, 3435--3444.

\bibitem[{Czolbe et~al.(2020)Czolbe, Krause, Cox, and Igel}]{czolbe2020loss}
Czolbe, S.; Krause, O.; Cox, I.; and Igel, C. 2020.
\newblock A loss function for generative neural networks based on watson’s perceptual model.
\newblock \emph{Advances in Neural Information Processing Systems}, 33: 2051--2061.

\bibitem[{Deng et~al.(2022)Deng, Tang, Dong, Ma, Pan, Wang, and Xu}]{deng2021stytr2}
Deng, Y.; Tang, F.; Dong, W.; Ma, C.; Pan, X.; Wang, L.; and Xu, C. 2022.
\newblock StyTr$^2$: Image Style Transfer with Transformers.
\newblock In \emph{IEEE Conference on Computer Vision and Pattern Recognition (CVPR)}.

\bibitem[{Dumoulin, Shlens, and Kudlur(2016)}]{dumoulin2016learned}
Dumoulin, V.; Shlens, J.; and Kudlur, M. 2016.
\newblock A learned representation for artistic style.
\newblock \emph{arXiv preprint arXiv:1610.07629}.

\bibitem[{Durall, Keuper, and Keuper(2020)}]{durall2020watch}
Durall, R.; Keuper, M.; and Keuper, J. 2020.
\newblock Watch your up-convolution: Cnn based generative deep neural networks are failing to reproduce spectral distributions.
\newblock In \emph{Proceedings of the IEEE/CVF conference on computer vision and pattern recognition}, 7890--7899.

\bibitem[{Durall, Pfreundt, and Keuper(2019)}]{durall2019stabilizing}
Durall, R.; Pfreundt, F.-J.; and Keuper, J. 2019.
\newblock Stabilizing GANs with Soft Octave Convolutions.
\newblock \emph{arXiv preprint arXiv:1905.12534}.

\bibitem[{Gatys, Ecker, and Bethge(2016)}]{gatys2016image}
Gatys, L.~A.; Ecker, A.~S.; and Bethge, M. 2016.
\newblock Image style transfer using convolutional neural networks.
\newblock In \emph{Proceedings of the IEEE conference on computer vision and pattern recognition}, 2414--2423.

\bibitem[{Gatys et~al.(2017)Gatys, Ecker, Bethge, Hertzmann, and Shechtman}]{gatys2017controlling}
Gatys, L.~A.; Ecker, A.~S.; Bethge, M.; Hertzmann, A.; and Shechtman, E. 2017.
\newblock Controlling perceptual factors in neural style transfer.
\newblock In \emph{Proceedings of the IEEE conference on computer vision and pattern recognition}, 3985--3993.

\bibitem[{Gentleman and Sande(1966)}]{gentleman1966fast}
Gentleman, W.~M.; and Sande, G. 1966.
\newblock Fast Fourier transforms: for fun and profit.
\newblock In \emph{Proceedings of the November 7-10, 1966, fall joint computer conference}, 563--578.

\bibitem[{Ghiasi et~al.(2017)Ghiasi, Lee, Kudlur, Dumoulin, and Shlens}]{ghiasi2017exploring}
Ghiasi, G.; Lee, H.; Kudlur, M.; Dumoulin, V.; and Shlens, J. 2017.
\newblock Exploring the structure of a real-time, arbitrary neural artistic stylization network.
\newblock \emph{arXiv preprint arXiv:1705.06830}.

\bibitem[{Howard et~al.(2017)Howard, Zhu, Chen, Kalenichenko, Wang, Weyand, Andreetto, and Adam}]{howard2017mobilenets}
Howard, A.~G.; Zhu, M.; Chen, B.; Kalenichenko, D.; Wang, W.; Weyand, T.; Andreetto, M.; and Adam, H. 2017.
\newblock Mobilenets: Efficient convolutional neural networks for mobile vision applications.
\newblock \emph{arXiv preprint arXiv:1704.04861}.

\bibitem[{Huang et~al.(2017)Huang, Chen, Li, Wu, Van Der~Maaten, and Weinberger}]{huang2017multi}
Huang, G.; Chen, D.; Li, T.; Wu, F.; Van Der~Maaten, L.; and Weinberger, K.~Q. 2017.
\newblock Multi-scale dense networks for resource efficient image classification.
\newblock \emph{arXiv preprint arXiv:1703.09844}.

\bibitem[{Huang and Belongie(2017)}]{huang2017arbitrary}
Huang, X.; and Belongie, S. 2017.
\newblock Arbitrary style transfer in real-time with adaptive instance normalization.
\newblock In \emph{Proceedings of the IEEE international conference on computer vision}, 1501--1510.

\bibitem[{Huo, Li, and Zhu(2021)}]{huo2021efficient}
Huo, F.; Li, B.; and Zhu, X. 2021.
\newblock Efficient wavelet boost learning-based multi-stage progressive refinement network for underwater image enhancement.
\newblock In \emph{Proceedings of the IEEE/CVF International Conference on Computer Vision}, 1944--1952.

\bibitem[{Jing et~al.(2020)Jing, Liu, Ding, Wang, Ding, Song, and Wen}]{jing2020dynamic}
Jing, Y.; Liu, X.; Ding, Y.; Wang, X.; Ding, E.; Song, M.; and Wen, S. 2020.
\newblock Dynamic Instance Normalization for Arbitrary Style Transfer.
\newblock In \emph{AAAI}.

\bibitem[{Johnson, Alahi, and Fei-Fei(2016)}]{johnson2016perceptual}
Johnson, J.; Alahi, A.; and Fei-Fei, L. 2016.
\newblock Perceptual losses for real-time style transfer and super-resolution.
\newblock In \emph{Computer Vision--ECCV 2016: 14th European Conference, Amsterdam, The Netherlands, October 11-14, 2016, Proceedings, Part II 14}, 694--711. Springer.

\bibitem[{Johnson and Frigo(2006)}]{johnson2006modified}
Johnson, S.~G.; and Frigo, M. 2006.
\newblock A modified split-radix FFT with fewer arithmetic operations.
\newblock \emph{IEEE Transactions on Signal Processing}, 55(1): 111--119.

\bibitem[{Ke, Maire, and Yu(2017)}]{ke2017multigrid}
Ke, T.-W.; Maire, M.; and Yu, S.~X. 2017.
\newblock Multigrid neural architectures.
\newblock In \emph{Proceedings of the IEEE Conference on Computer Vision and Pattern Recognition}, 6665--6673.

\bibitem[{Kingma and Ba(2014)}]{kingma2014adam}
Kingma, D.~P.; and Ba, J. 2014.
\newblock Adam: A method for stochastic optimization.
\newblock \emph{arXiv preprint arXiv:1412.6980}.

\bibitem[{Kotovenko et~al.(2019)Kotovenko, Sanakoyeu, Lang, and Ommer}]{kotovenko2019content}
Kotovenko, D.; Sanakoyeu, A.; Lang, S.; and Ommer, B. 2019.
\newblock Content and style disentanglement for artistic style transfer.
\newblock In \emph{Proceedings of the IEEE/CVF international conference on computer vision}, 4422--4431.

\bibitem[{Li et~al.(2020)Li, Cai, Li, Cao, Wang, and Li}]{li2020frequency}
Li, S.; Cai, Q.; Li, H.; Cao, J.; Wang, L.; and Li, Z. 2020.
\newblock Frequency separation network for image super-resolution.
\newblock \emph{IEEE Access}, 8: 33768--33777.

\bibitem[{Li et~al.(2017)Li, Fang, Yang, Wang, Lu, and Yang}]{li2017universal}
Li, Y.; Fang, C.; Yang, J.; Wang, Z.; Lu, X.; and Yang, M.-H. 2017.
\newblock Universal style transfer via feature transforms.
\newblock \emph{Advances in neural information processing systems}, 30.

\bibitem[{Lin et~al.(2017)Lin, Doll{\'a}r, Girshick, He, Hariharan, and Belongie}]{lin2017feature}
Lin, T.-Y.; Doll{\'a}r, P.; Girshick, R.; He, K.; Hariharan, B.; and Belongie, S. 2017.
\newblock Feature pyramid networks for object detection.
\newblock In \emph{Proceedings of the IEEE conference on computer vision and pattern recognition}, 2117--2125.

\bibitem[{Lin et~al.(2014)Lin, Maire, Belongie, Hays, Perona, Ramanan, Doll{\'a}r, and Zitnick}]{lin2014microsoft}
Lin, T.-Y.; Maire, M.; Belongie, S.; Hays, J.; Perona, P.; Ramanan, D.; Doll{\'a}r, P.; and Zitnick, C.~L. 2014.
\newblock Microsoft coco: Common objects in context.
\newblock In \emph{Computer Vision--ECCV 2014: 13th European Conference, Zurich, Switzerland, September 6-12, 2014, Proceedings, Part V 13}, 740--755. Springer.

\bibitem[{Lindeberg(2013)}]{lindeberg2013scale}
Lindeberg, T. 2013.
\newblock \emph{Scale-space theory in computer vision}, volume 256.
\newblock Springer Science \& Business Media.

\bibitem[{Liu et~al.(2021{\natexlab{a}})Liu, Lin, He, Li, Wang, Li, Sun, Li, and Ding}]{liu2021adaattn}
Liu, S.; Lin, T.; He, D.; Li, F.; Wang, M.; Li, X.; Sun, Z.; Li, Q.; and Ding, E. 2021{\natexlab{a}}.
\newblock Adaattn: Revisit attention mechanism in arbitrary neural style transfer.
\newblock In \emph{Proceedings of the IEEE/CVF international conference on computer vision}, 6649--6658.

\bibitem[{Liu et~al.(2021{\natexlab{b}})Liu, Meng, Tan, Zhang, and Zhang}]{liu2021image}
Liu, Z.; Meng, L.; Tan, Y.; Zhang, J.; and Zhang, H. 2021{\natexlab{b}}.
\newblock Image compression based on octave convolution and semantic segmentation.
\newblock \emph{Knowledge-Based Systems}, 228: 107254.

\bibitem[{Lowe(2004)}]{lowe2004distinctive}
Lowe, D.~G. 2004.
\newblock Distinctive image features from scale-invariant keypoints.
\newblock \emph{International journal of computer vision}, 60: 91--110.

\bibitem[{Paszke et~al.(2019)Paszke, Gross, Massa, Lerer, Bradbury, Chanan, Killeen, Lin, Gimelshein, Antiga et~al.}]{paszke2019pytorch}
Paszke, A.; Gross, S.; Massa, F.; Lerer, A.; Bradbury, J.; Chanan, G.; Killeen, T.; Lin, Z.; Gimelshein, N.; Antiga, L.; et~al. 2019.
\newblock Pytorch: An imperative style, high-performance deep learning library.
\newblock \emph{Advances in neural information processing systems}, 32.

\bibitem[{Phillips and Mackintosh(2011)}]{phillips2011wiki}
Phillips, F.; and Mackintosh, B. 2011.
\newblock Wiki Art Gallery, Inc.: A case for critical thinking.
\newblock \emph{Issues in Accounting Education}, 26(3): 593--608.

\bibitem[{Robinson et~al.(2020)Robinson, Chuang, Sra, and Jegelka}]{robinson2020contrastive}
Robinson, J.; Chuang, C.-Y.; Sra, S.; and Jegelka, S. 2020.
\newblock Contrastive learning with hard negative samples.
\newblock \emph{arXiv preprint arXiv:2010.04592}.

\bibitem[{Shen, Yan, and Zeng(2018)}]{shen2018neural}
Shen, F.; Yan, S.; and Zeng, G. 2018.
\newblock Neural style transfer via meta networks.
\newblock In \emph{Proceedings of the IEEE Conference on Computer Vision and Pattern Recognition}, 8061--8069.

\bibitem[{Sheng et~al.(2018)Sheng, Lin, Shao, and Wang}]{sheng2018avatar}
Sheng, L.; Lin, Z.; Shao, J.; and Wang, X. 2018.
\newblock Avatar-net: Multi-scale zero-shot style transfer by feature decoration.
\newblock In \emph{Proceedings of the IEEE conference on computer vision and pattern recognition}, 8242--8250.

\bibitem[{Simonyan and Zisserman(2014)}]{simonyan2014very}
Simonyan, K.; and Zisserman, A. 2014.
\newblock Very deep convolutional networks for large-scale image recognition.
\newblock \emph{arXiv preprint arXiv:1409.1556}.

\bibitem[{Sun et~al.(2019)Sun, Xiao, Liu, and Wang}]{sun2019deep}
Sun, K.; Xiao, B.; Liu, D.; and Wang, J. 2019.
\newblock Deep high-resolution representation learning for human pose estimation.
\newblock In \emph{Proceedings of the IEEE/CVF conference on computer vision and pattern recognition}, 5693--5703.

\bibitem[{Szegedy et~al.(2015)Szegedy, Liu, Jia, Sermanet, Reed, Anguelov, Erhan, Vanhoucke, and Rabinovich}]{szegedy2015going}
Szegedy, C.; Liu, W.; Jia, Y.; Sermanet, P.; Reed, S.; Anguelov, D.; Erhan, D.; Vanhoucke, V.; and Rabinovich, A. 2015.
\newblock Going deeper with convolutions.
\newblock In \emph{Proceedings of the IEEE conference on computer vision and pattern recognition}, 1--9.

\bibitem[{Ulyanov et~al.(2016)Ulyanov, Lebedev, Vedaldi, and Lempitsky}]{ulyanov2016texture}
Ulyanov, D.; Lebedev, V.; Vedaldi, A.; and Lempitsky, V. 2016.
\newblock Texture networks: Feed-forward synthesis of textures and stylized images.
\newblock \emph{arXiv preprint arXiv:1603.03417}.

\bibitem[{Ulyanov, Vedaldi, and Lempitsky(2017)}]{ulyanov2017improved}
Ulyanov, D.; Vedaldi, A.; and Lempitsky, V. 2017.
\newblock Improved texture networks: Maximizing quality and diversity in feed-forward stylization and texture synthesis.
\newblock In \emph{Proceedings of the IEEE conference on computer vision and pattern recognition}, 6924--6932.

\bibitem[{Van~Loan(1992)}]{van1992computational}
Van~Loan, C. 1992.
\newblock \emph{Computational frameworks for the fast Fourier transform}.
\newblock SIAM.

\bibitem[{Wang et~al.(2019)Wang, Kembhavi, Farhadi, Yuille, and Rastegari}]{wang2019elastic}
Wang, H.; Kembhavi, A.; Farhadi, A.; Yuille, A.~L.; and Rastegari, M. 2019.
\newblock Elastic: Improving cnns with dynamic scaling policies.
\newblock In \emph{Proceedings of the IEEE/CVF Conference on Computer Vision and Pattern Recognition}, 2258--2267.

\bibitem[{Wang et~al.(2020{\natexlab{a}})Wang, Wu, Huang, and Xing}]{wang2020high}
Wang, H.; Wu, X.; Huang, Z.; and Xing, E.~P. 2020{\natexlab{a}}.
\newblock High-frequency component helps explain the generalization of convolutional neural networks.
\newblock In \emph{Proceedings of the IEEE/CVF conference on computer vision and pattern recognition}, 8684--8694.

\bibitem[{Wang et~al.(2021)Wang, Guo, Huang, Li, Zhang, and Jiao}]{wang2021dual}
Wang, J.; Guo, S.; Huang, R.; Li, L.; Zhang, X.; and Jiao, L. 2021.
\newblock Dual-channel capsule generation adversarial network for hyperspectral image classification.
\newblock \emph{IEEE Transactions on Geoscience and Remote Sensing}, 60: 1--16.

\bibitem[{Wang, Li, and Vasconcelos(2021)}]{wang2021rethinking}
Wang, P.; Li, Y.; and Vasconcelos, N. 2021.
\newblock Rethinking and improving the robustness of image style transfer.
\newblock In \emph{Proceedings of the IEEE/CVF Conference on Computer Vision and Pattern Recognition}, 124--133.

\bibitem[{Wang et~al.(2020{\natexlab{b}})Wang, Khan, Gonzalez-Garcia, Weijer, and Khan}]{wang2020semi}
Wang, Y.; Khan, S.; Gonzalez-Garcia, A.; Weijer, J. v.~d.; and Khan, F.~S. 2020{\natexlab{b}}.
\newblock Semi-supervised learning for few-shot image-to-image translation.
\newblock In \emph{Proceedings of the IEEE/CVF Conference on Computer Vision and Pattern Recognition}, 4453--4462.

\bibitem[{Wang et~al.(2004)Wang, Bovik, Sheikh, and Simoncelli}]{wang2004image}
Wang, Z.; Bovik, A.~C.; Sheikh, H.~R.; and Simoncelli, E.~P. 2004.
\newblock Image quality assessment: from error visibility to structural similarity.
\newblock \emph{IEEE transactions on image processing}, 13(4): 600--612.

\bibitem[{Wang et~al.(2022)Wang, Zhang, Zhao, Zuo, Li, Xing, and Lu}]{wang2022aesust}
Wang, Z.; Zhang, Z.; Zhao, L.; Zuo, Z.; Li, A.; Xing, W.; and Lu, D. 2022.
\newblock AesUST: towards aesthetic-enhanced universal style transfer.
\newblock In \emph{Proceedings of the 30th ACM International Conference on Multimedia}, 1095--1106.

\bibitem[{Wang et~al.(2023)Wang, Zhao, Zuo, Li, Chen, Xing, and Lu}]{wang2023microast}
Wang, Z.; Zhao, L.; Zuo, Z.; Li, A.; Chen, H.; Xing, W.; and Lu, D. 2023.
\newblock MicroAST: Towards Super-Fast Ultra-Resolution Arbitrary Style Transfer.
\newblock In \emph{Proceedings of the AAAI Conference on Artificial Intelligence}.

\bibitem[{Xu et~al.(2020)Xu, Qin, Sun, Wang, Chen, and Ren}]{xu2020learning}
Xu, K.; Qin, M.; Sun, F.; Wang, Y.; Chen, Y.-K.; and Ren, F. 2020.
\newblock Learning in the frequency domain.
\newblock In \emph{Proceedings of the IEEE/CVF Conference on Computer Vision and Pattern Recognition}, 1740--1749.

\bibitem[{Xu et~al.(2019)Xu, Zhang, Luo, Xiao, and Ma}]{xu2019frequency}
Xu, Z.-Q.~J.; Zhang, Y.; Luo, T.; Xiao, Y.; and Ma, Z. 2019.
\newblock Frequency principle: Fourier analysis sheds light on deep neural networks.
\newblock \emph{arXiv preprint arXiv:1901.06523}.

\bibitem[{Zhang et~al.(2018)Zhang, Isola, Efros, Shechtman, and Wang}]{zhang2018unreasonable}
Zhang, R.; Isola, P.; Efros, A.~A.; Shechtman, E.; and Wang, O. 2018.
\newblock The unreasonable effectiveness of deep features as a perceptual metric.
\newblock In \emph{Proceedings of the IEEE conference on computer vision and pattern recognition}, 586--595.

\bibitem[{Zhang et~al.(2022{\natexlab{a}})Zhang, Li, Li, Jia, and Zhang}]{zhang2022exact}
Zhang, Y.; Li, M.; Li, R.; Jia, K.; and Zhang, L. 2022{\natexlab{a}}.
\newblock Exact feature distribution matching for arbitrary style transfer and domain generalization.
\newblock In \emph{Proceedings of the IEEE/CVF Conference on Computer Vision and Pattern Recognition}, 8035--8045.

\bibitem[{Zhang et~al.(2022{\natexlab{b}})Zhang, Li, Qi, Liu, Kong, and Wang}]{zhang2022multi}
Zhang, Y.; Li, Q.; Qi, M.; Liu, D.; Kong, J.; and Wang, J. 2022{\natexlab{b}}.
\newblock Multi-scale frequency separation network for image deblurring.
\newblock \emph{arXiv preprint arXiv:2206.00798}.

\bibitem[{Zhao et~al.(2017)Zhao, Shi, Qi, Wang, and Jia}]{zhao2017pyramid}
Zhao, H.; Shi, J.; Qi, X.; Wang, X.; and Jia, J. 2017.
\newblock Pyramid scene parsing network.
\newblock In \emph{Proceedings of the IEEE conference on computer vision and pattern recognition}, 2881--2890.

\end{thebibliography}
\end{document}